\pdfoutput=1

\documentclass[11pt]{article}
\usepackage{algorithm}
\usepackage{algpseudocode}
\usepackage[final]{acl}
\usepackage{booktabs} 
\usepackage{multirow} 

\usepackage{times}
\usepackage{graphicx}
\usepackage{latexsym}
\usepackage{amsmath}
\usepackage[T1]{fontenc}

\usepackage[utf8]{inputenc}
\usepackage{caption,booktabs}

\usepackage{microtype}
\usepackage{xcolor}
\usepackage{inconsolata}

\usepackage{graphicx}
\usepackage{stackengine}
\usepackage{scalerel}
\usepackage{xcolor,amssymb}

\newcommand\dangersignw[1][2ex]{%
  \scaleto{\stackengine{0.3pt}{\scalebox{1.1}[.9]{%
  \color{red}$\blacktriangle$}}{\color{white}\tiny\bfseries !}{O}{c}{F}{F}{L}}{#1}%
}
\title{ToxiCraft: A Novel Framework for Synthetic Generation of Harmful Information}

\author{
 \textbf{Zheng Hui\textsuperscript{$\clubsuit$ $\heartsuit$}},
 \textbf{Zhaoxiao Guo\textsuperscript{$\clubsuit$ $\dagger$}},
 \textbf{Hang Zhao\textsuperscript{$\clubsuit$}},
 \textbf{Juanyong Duan\textsuperscript{$\clubsuit$}},
 \textbf{Congrui Huang\textsuperscript{$\clubsuit$}}
\\
\\
 \textsuperscript{$\clubsuit$}Microsoft,
 \textsuperscript{$\heartsuit$}Columbia University,
  \textsuperscript{$\dagger$}Tsinghua University
\\
    \{zackhui, hang.zhao, juanyong.duan, conhua\}@microsoft.com,    
    \\
    zh2483@columbia.edu, guozx22@mails.tsinghua.edu.cn
\\
 \small{
   \textbf{Correspondence:} \href{mailto:juaduan@microsoft.com}{juaduan@microsoft.com}
 }
}

\begin{document}
\maketitle
\begin{abstract}
In different NLP tasks, detecting harmful content is crucial for online environments, especially with the growing influence of social media. However, previous research has two main issues: 1) a lack of data in low-resource settings, and 2) inconsistent definitions and criteria for judging harmful content, requiring classification models to be robust to spurious features and diverse. We propose \textbf{ToxiCraft}, a novel framework for synthesizing datasets of harmful information to address these weaknesses. With only a small amount of seed data, our framework can generate a wide variety of synthetic, yet remarkably realistic, examples of toxic information. Experimentation across various datasets showcases a notable enhancement in detection model robustness and adaptability, surpassing or close to the gold labels. Dataset is available at \href{https://github.com/zackhuiiiii/ToxiCraft}{this GitHub repository.}

\color{red} 
\dangersignw This paper has instances of hateful and offensive language to serve as examples.
\end{abstract}

\section{Introduction}
The 21st-century digital realm presents vast connectivity and information exchange opportunities alongside the challenge of widespread harmful content like cyberbullying, hate speech, and misinformation, impacting individuals and communities negatively. As such, the development of effective mechanisms for detecting and mitigating harmful content is of paramount importance \cite{breitfeller-etal-2019-finding,casula-tonelli-2023-generation,plaza-del-arco-etal-2023-respectful}.

\begin{figure}[t]

  \centerline{\includegraphics[width=0.49\textwidth]{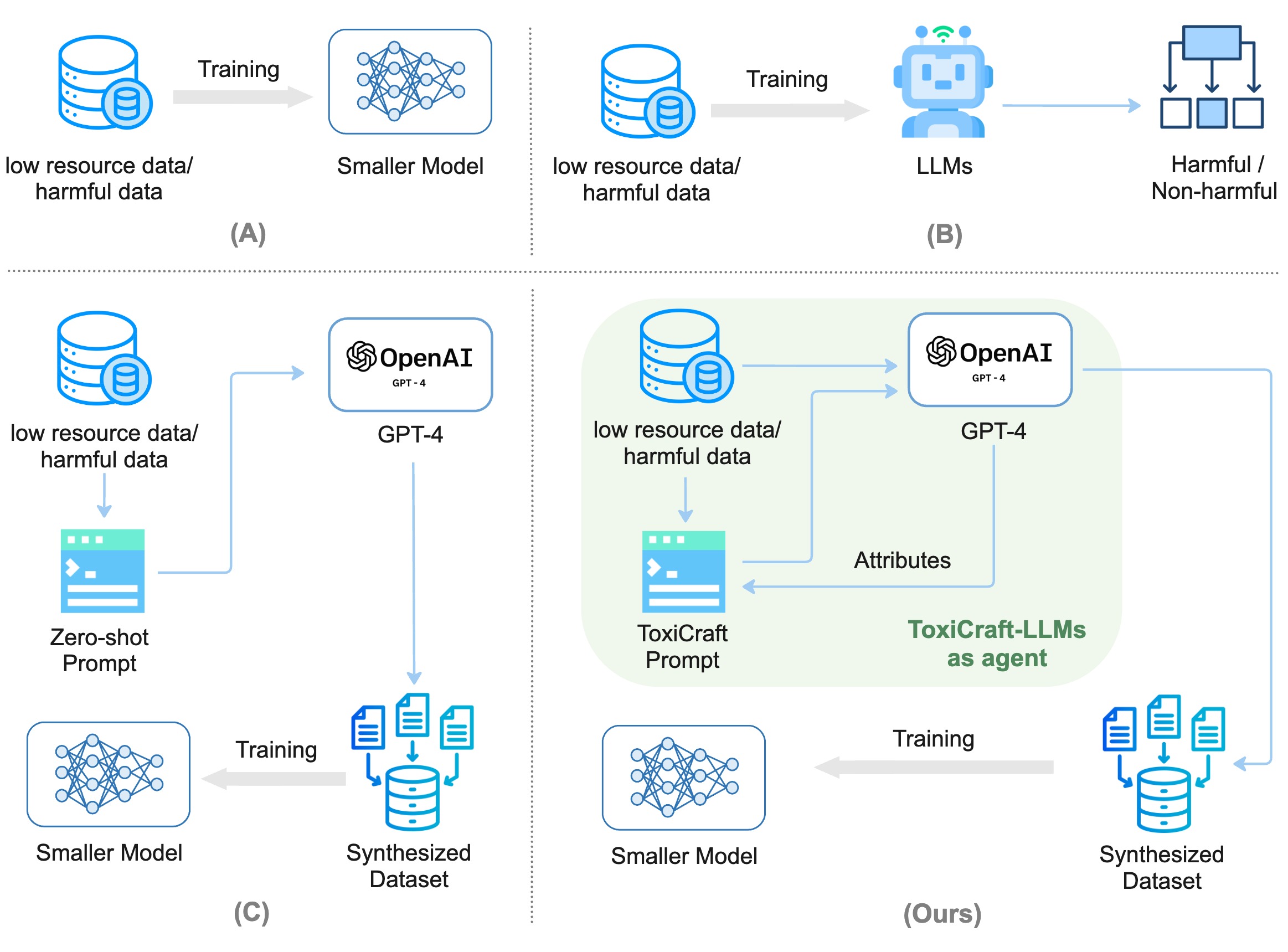}}
  \captionsetup{justification=centering}
  \caption{Harmful detection approaches}
  \label{fig:f1}
\end{figure}

The emergence of the Transformer-based model \cite{NIPS2017_3f5ee243} led to the development of complex models capable of identifying toxic content with remarkable accuracy. However, the effectiveness of these models hinges on the quality and diversity of the datasets used for their training \cite{banko2001mitigating}. Traditional datasets, often curated manually, tend to lack the diversity required to cover the multifaceted nature of harmful content. Consequently, while these models excel at recognizing explicit instances of toxicity, they struggle when faced with subtler forms \cite{casula-tonelli-2023-generation}.

Many datasets containing harmful content are typically sourced from social media platforms like Twitter or online forums \cite{davidson2017automated, grimminger2021hate, waseem-hovy-2016-hateful, sachdeva-etal-2022-measuring}. However, these datasets often exhibit significant class imbalances, particularly concerning specific types of toxic language. Training a model to detect specific harmful content or opting for smaller models due to resource or latency constraints requires substantial amounts of human-labeled data. Unfortunately, such data are scarce in downstream tasks and expensive to annotate \cite{breitfeller-etal-2019-finding, juuti-etal-2020-little}. Additionally, smaller or low-resource datasets face challenges like limited linguistic diversity and a higher risk of overfitting \cite{klubička2018examining}. Privacy concerns also emerge when using social media data obtained without user consent for research purposes \cite{casula-tonelli-2023-generation}. Furthermore, dataset decay is also a notable issue, especially with online social media, including abusive content, which are often deleted over time \cite{casula-tonelli-2023-generation}\footnote{Twitter's policy only allows the display of post IDs rather than the text itself, making content inaccessible over time, often due to deletion by either the user or Twitter, particularly in cases involving harmful information.}.

Recent advancements in large language models (LLMs) such as the GPT series \cite{brown2020language} have prompted researchers to explore their potential for generating synthetic data \cite{yoo2021gpt3mix,ye2022zerogen,gao2023selfguided}. While LLM-based data augmentation often improves model performance, findings are mixed regarding whether synthetic data generated by LLMs can consistently match the effectiveness of models trained on real-world, human-annotated data \cite{li-etal-2023-synthetic}.
LLMs have also shown promise in generating synthetic data for training harmful content detection models. However, work in this area remains limited \cite{sen-etal-2023-people} and much of the existing research is relatively outdated \cite{juuti-etal-2020-little,geetdsa:hal-03244472,liu-etal-2020-data,geetdsa:hal-03244472,wullach-etal-2021-fight-fire}. Moreover, the effectiveness of LLMs in this domain is hindered by their difficulty with subjective tasks and their tendency to amplify biases inherent in human-annotated training data. These biases stem from inconsistent standards used in defining `gold' labels, and LLMs often struggle to capture the complex nuances and diversity of human communication in highly subjective tasks.

To address these challenges, we introduce the ToxiCraft Framework, which enhances robustness and reduces bias in synthetic data generation. The ToxiCraft Framework is specifically designed to produce higher-quality synthetic data that better supports model training. An example of the process is illustrated in Figure \ref{fig:f1}. In Figure \ref{fig:f1}, (A) is direct training where a smaller model is trained on a limited dataset, that may affect model quality. (B) shows the use of large language models to do harmful detaction. (C) illustrates Zero-shot Synthesis, where use LLMs to generate a synthetic dataset via zero-shot prompts, used to train smaller models without additional data. Our ToxiCraft method (D) employs carefully crafted prompts and attributes, utilizing GPT-4 as the backbone to generate a synthetic dataset for training more refined smaller models. A more detailed explanation of our approach can be found in Section \ref{method}. To summarise, our work has the following key contributions:

\begin{itemize}
  \item We identify a limitation of traditional simple zero-shot prompting methods in data synthesis which leads to a lack of diversity.
  
  \item Enhances low-resource harmful information datasets to address the pervasive issue of harmful dataset sources decay, making the enhanced datasets publicly available to ensure reproducibility and broader utility.
  
  \item We introduce a novel data synthetic framework, mitigating biases and ensuring a closer reflection of real-world complexities and demonstrating its efficacy in generating the diverse and high-quality data. 
  
  \item Applications of the synthesized data in fine-tuning smaller models, highlight the practical value of LLM generated data.
  
\end{itemize}

\section{Related Work}

\subsection{Harmful Information Detection} 
 Early studies established classifiers to detect harmful information using neural network models \cite{Zhang2018DetectingHS} or word embedding methods \cite{kshirsagar2018predictive}. In recent years, models based on the Transformer architecture have demonstrated remarkable capabilities, prompting researchers to explore further. \citet{Rajput_2021} conducted research on the ETHOS hate speech detection dataset, comparing classifiers' performance in hate speech detection by replacing or integrating word embeddings (fastText, GloVe, or FT + GV) with BERT embeddings. \citet{aluru2020deep} contrasted simple models (such as LASER embeddings with logistic regression) and BERT models in scenarios with scarce and abundant linguistic resources. \citet{lin2024explainable} generated explanations through multimodal debates between LLMs, enhancing the transparency and explainability of harmful meme detection. \citet{da-silva-oliveira-etal-2024-toxic} validated the effectiveness of ChatGPT in identifying harmful Spanish-language speech. Our ToxiCraft framework extends these developments by leveraging LLMs not only for direct detection but also for enhancement of the training data pool through synthetic data generation. Unlike methods that rely solely on existing data, ToxiCraft enriches the dataset, which is particularly critical in the ever-evolving domain of online content where new forms of harmful expression continuously emerge.

\subsection{Large Language Models} 
Transformer \cite{vaswani2023attention} architecture has revolutionized the field of NLP \cite{chang2023survey}. Large Language Models (LLMs) based on the Transformer architecture are pre-trained on extensive corpora, resulting in models with vast parameter scales and exceptional learning capabilities. The BERT model \cite{devlin-etal-2019-bert} utilizes the bidirectional encoder from Transformer to process input text, generating rich context-aware word embeddings, with variants including ALBERT \cite{lan2020albert}, RoBERTa\cite{liu2019roberta}, and DeBERTa\cite{he2021deberta}. OpenAI have launched the GPT series of large language models based on Transformer's decoder, including GPT-2 \cite{Radford2019LanguageMA}, GPT-3 \cite{brown2020language}, InstructGPT \cite{ouyang2022training}, and GPT-4 \cite{openai2024gpt4}. These large language models are garnering increasing research interest, not only exhibiting outstanding performance in a broad range of tasks in the natural language understanding sector—such as sentiment analysis \cite{scaria2023instructabsa,fei-etal-2023-reasoning}, text classification \cite{Pe_a_2023} but also demonstrating remarkable capabilities in natural language generation tasks like summarization \cite{wang-etal-2023-element}, translation \cite{wang-etal-2023-document-level}, and question answering \cite{yan-etal-2021-large}.

\subsection{Data Synthesis} 
Traditional data synthesis methods have employed techniques ranging from synonym replacement to token-level manipulations, as exemplified by the works of \cite{zhang2016characterlevel} and \cite{wei-zou-2019-eda}. These methods, while useful, offer limited contextual understanding and diversity. The advent of translation models \cite{Fadaee_2017} and masked filling \cite{kumar-etal-2020-data} brought improvements in maintaining semantic consistency, yet they still fall short in generating the context-rich synthetic data required for complex tasks such as harmful content detection.
In contrast, ToxiCraft leverages the latest advancements in LLMs for data synthesis, transcending the limitations of earlier approaches by producing contextually nuanced and varied synthetic instances. Unlike methods that require fine-tuning on annotated data, which incurs substantial human labor costs \cite{yang-etal-2020-generative, mohapatra2021simulated, kumar2021data}, ToxiCraft efficiently synthesizes data without intensive manual effort. In comparison to zero-shot frameworks like ZEROGEN \cite{ye2022zerogen}, SuperGen \cite{meng2022generating}, and PROGEN \cite{ye2022progen}, which generate datasets from scratch or suffer from low information content and redundancy, ToxiCraft’s approach is designed to yield high-quality synthetic data that is both diverse and relevant to the seed data, thus ensuring its applicability to real-world tasks.
In the few-shot learning domain, while \citet{yu2023large} rely on diverse attribute prompts and PromDA \cite{wang-etal-2022-promda} along with MSP \cite{chen2023mixture} utilize soft prompts for data diversity and optimization, ToxiCraft differentiates itself by not only enhancing the diversity but also focusing on the generation of synthetic data that closely aligns with the intricacies of harmful content. The integration of attribute prompts within ToxiCraft’s framework ensures that the synthesized data captures a broad spectrum of harmful content, effectively addressing both the volume and variety required for robust model training.

\begin{figure*}[ht]
  \begin{center}
  \includegraphics[width=2.08\columnwidth]{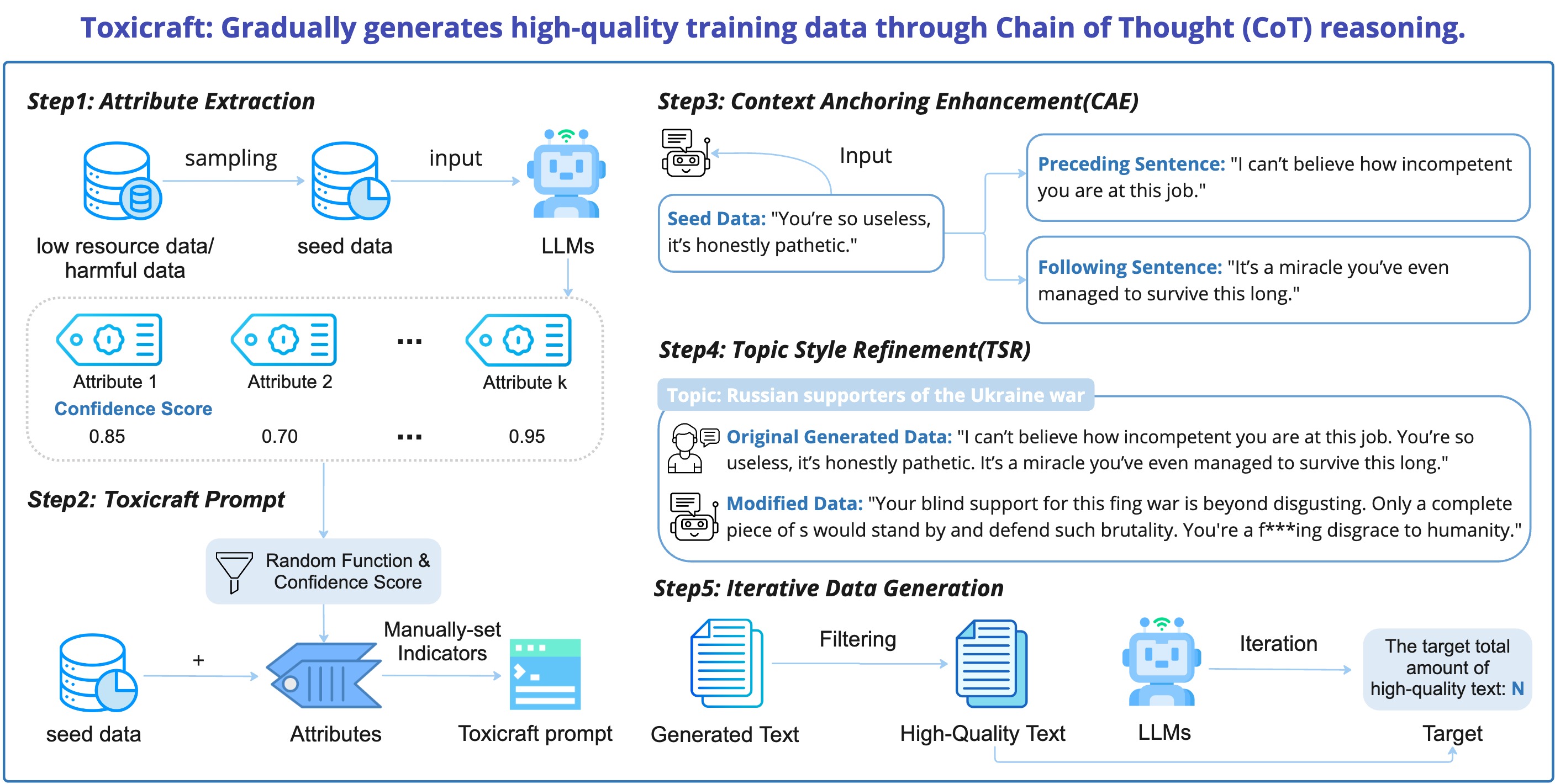}
  \end{center}
  \captionsetup{justification=centering}
  \caption{Our \textbf{ToxiCraft} Framework Diagram, CAE \& TSR step is elaborating on transforming the initial data into multiple versions by altering the context while maintaining the core message or change the topic}
  \label{fig:f2}
\end{figure*}

\section{Methodology}\label{method}

In this section, we commence by defining the problem statement and subsequently introduce the ToxiCraft framework. ToxiCraft is designed as a versatile and effective few-shot learning framework that operates through dataset generation. ToxiCraft framework are illustrated in Figure \ref{fig:f2}.
\subsection{Problem Statement }

In the context of online content moderation, particularly the detection of harmful information, the problem can be modeled as a supervised classification task. Given an input content sequence \(x\), represented as tokens \(x = [x_1, x_2, \ldots, x_{|x|}]\), the goal is to predict a binary output \(y\), where \(y \in \{0, 1\}\) indicates whether the content is harmful (\(y=1\)) or not (\(y=0\)). The dataset \(D\) consists of pairs of content sequences and their corresponding labels:
\[
D = \{(x^{(i)}, y^{(i)})\}_{i=1}^N
\]
where \(N\) represents the number of examples in the dataset. The major challenge in this domain is the scarcity of labeled examples of harmful content, denoted as \(D_h\):
\[
D_h = \{(x^{(i)}, y^{(i)}) \in D : y^{(i)} = 1\}
\]
Due to the sparse nature of \(D_h\), there is a need for effective methods to augment the dataset, especially enhancing the representation of the minority class, which is harmful content in this scenario.

To address the limitations posed by scarce harmful content samples, we propose a data augmentation method - \textbf{ToxiCraft} that involves synthesizing new content instances using LLMs. ToxiCraft utilizes LLMs to augment the dataset with synthetic instances, enhancing the representational diversity of harmful content. The data augmentation function, defined by:
\[
\text{Aug}(x) = \{x' : x' = \Phi(\text{LLM}, x, \epsilon), \epsilon \sim \mathcal{P}(\epsilon)\}
\]
Here, \( \Phi \) denotes the ToxiCraft framework which is described in Section \ref{toxicraft}. And ToxiCraft uses selected seed data to guide the synthesis, ensuring that the generated content is relevant and varied.
\(\epsilon\) represents a noise vector drawn from a probability distribution \(\mathcal{P}\), introducing variations in the synthesized outputs to prevent overfitting to the seed samples. Seed data \(S\) for ToxiCraft are selected from \(D_h\) based on a random sampling process, which aims to capture a diverse array of harmful content types. The selection function \(S(d, n)\) picks \(n\) samples from \(d\) using a probability distribution tailored to emphasize less represented content:
\[
\begin{aligned}
S(D_h, n) &= \{(x^{(i)}, y^{(i)}) : x^{(i)} \in D_h, \\
&\quad i \in \text{random indices from } D_h\}
\end{aligned}
\]

\noindent where \(n\) is the number of seed data chosen to guide the LLM in data generation. Details on the selection process and the impact of seed data size are provided in Section \ref{exp}. The augmented dataset \(D'\) is then:
\[
D' = D \cup \{(x', y) : x' \in \text{Aug}(x), S(D_h, n) \in D_h\}
\]

\noindent After that, we employ a smaller model, e.g., BERT, fine-tuned on both \(D\) and \(D'\). The training objective is to minimize the loss function \(\mathcal{L}(\theta)\) across the augmented dataset:
\[
\mathcal{L}(\theta) = \frac{1}{|D'|} \sum_{(x, y) \in D'} \ell(h_{\theta}(x), y)
\]
where \(\ell\) represents a loss function appropriate for binary classification tasks, such as cross-entropy.


\noindent The evaluation of the ToxiCraft framework using the smaller model involves several key metrics:

\textbf{Effectiveness:} Measured by the accuracy and Macro F1-score of the model on a held-out test set \(T\), not seen during training or augmentation:
\[
\text{Macro F1-Score}(\theta) = \frac{1}{K} \sum_{k=1}^{K} F1_k(\theta)
\]
where \(F1_k(\theta) = 2 \times \frac{\text{Precision}_k(\theta) \times \text{Recall}_k(\theta)}{\text{Precision}_k(\theta) + \text{Recall}_k(\theta)}\) and \(K\) is the number of classes, which in this binary classification context, \(K=2\).

\textbf{Robustness:} In our setting, robustness is evaluated using cross-dataset evaluation, where data augmented on one dataset is used to evaluate performance on other datasets. Specifically, robustness is measured by the model's performance consistency across different datasets \(\mathcal{D}_{i}\), with data augmented from one dataset \(\mathcal{D}_{j}\) and evaluated on another dataset \(\mathcal{D}_{k}\). The variance of the model's loss across these datasets captures the robustness:
\[
\text{Robustness}(\theta, \mathcal{D}) = \text{Var}_{\mathcal{D}_{k}}(\ell(h_{\theta}(\mathcal{D}_{k}), y_k))
\]
where \(h_{\theta}\) represents the model parameterized by \(\theta\), \(\ell(\cdot)\) is the loss function, and \(\mathcal{D}_{k}\) are the instances from different datasets. \(\text{Var}\) captures the variance in performance across these datasets, reflecting the robustness of the model when trained on data from \(\mathcal{D}_{j}\) and tested on others. We also provide a pseudocode in Appendix \ref{sec:appendix1}.

\subsection{ToxiCraft Framework}\label{toxicraft}

In the ToxiCraft framework, we employed LLMs as agents to analyze seed data containing harmful information. 

\noindent \textbf{Attribute Extraction}
The LLMs' task is to identify the attributes within this data to determine if the content is harmful. This concept was inspired by the research of \citet{yu2023large}; however, unlike their approach, we did not manually filter the attributes generated by GPT. Instead, we retained these attribute tags generated by the model itself and cataloged all specific groups or themes indicating harm in the seed data, collectively termed the `Harmful Themes Index'. The advantage of automated attribute tagging is its capacity to process and analyze data rapidly and on a large scale, significantly enhancing processing efficiency. To further enhance the accuracy of generated attributes and the robustness of the model, we introduced a new step where LLM provides a confidence score when generating attributes. Based on this confidence score, we decided whether to retain the attribute using a simple random function.

Even with a high confidence score, an attribute may be randomly discarded by a small chance, which could increase the robustness of the decision-making process, reduce the risk of cumulative errors, and avoid over-reliance on the model's singular judgment. In this process, we considered the possibility of errors in automated attribute tagging but decided not to take model-based corrective measures. Instead, we chose to maintain a simple and efficient system to quickly respond to and update changes in the dataset. Although this approach is simple, it can maintain a certain level of accuracy and consistency without sacrificing efficiency.

\noindent \textbf{ToxiCraft Prompt}
Next, we added manually-set indicators to this data with Attribute Extraction by GPT (we call this process as `ToxiCraft Prompt') and input 10\% of the data extracted randomly from the seed data pool together with `ToxiCraft Prompt' to GPT. Manually-set indicators include the intensification or weakening of tone, whether to increase swear words, whether to use irony, country and time. Each value will be randomly selected or masked. Thus, it is possible that none of the manually-set indicators are added, but at most only one answer will be selected from each indicator. For example, the country could be randomly set as the United States and the year as 2023.

\noindent \textbf{CAE and TSR}
According to the research by \citet{pavlopoulos-etal-2020-toxicity} and \cite{geetdsa:hal-03244472}, high-quality context can help better detect harmful data. Considering the inherent length limitations of the dataset (primarily from a harmful dataset on Twitter before 2017 with a limit of 140 characters (about 20-35 words) and after 2017 with a limit of 280 characters), we innovatively generated context-based preceding and succeeding text. We employed a technique called `Contextual Anchoring Enhancement', using dropout to randomly abandon the preceding or succeeding text or retain them all to enhance the model's robustness. Furthermore, we also checked the generated data and used GPT to evaluate their quality. Among the result data, we selected the top-performing 10\%, applied `Thematic Style Refinement', transforming the themes recorded in the `Harmful Themes Index', and added them to the seed data. We repeated these steps until generating $M$ results controlled by hyperparameters, while the number of data generated per batch using 10\% of seed data $K$ was also controlled by hyperparameters. The total ToxiCraft framework genrated data are $N$.

ToxiCraft effectively utilizes the Chain of Thought (COT) concept to gradually generate high-quality training data. Additionally, we compared the dataset generated by ToxiCraft with the dataset generated only using simple COT prompts, and detailed experimental results will be presented in Section \ref{exp}.

\section{Experiment Setup}\label{exp}

\subsection{Datasets and Processing}
In this section, we outline the data employed in our experiments. We first introduce the four datasets containing harmful information in Section \ref{datasets}. Following this, we consolidate these datasets and present statistics in Section \ref{dataprocessing}. 

\subsubsection{Harmful Information Datasets}\label{datasets}

\textbf{Automated Hate Speech Detection (AHSD)} described by \citet{davidson2017automated}, comprises 24,802 tweets extracted using a hate speech lexicon. These tweets are manually categorized by CrowdFlower workers into three distinct groups: hate speech, offensive but not hate speech, and neither, with an intercoder agreement of 92\%.

\noindent \textbf{Towards the Political Opponent (HTPO)} described by \citet{grimminger-klinger-2021-hate} encompasses  3,000 tweets related to the 2020 U.S. Presidential election, annotated for both stance detection (favorable or against candidates) and the presence of hateful language, enabling nuanced analysis of political discourse and sentiment.

\noindent \textbf{Hate Speech and Offensive Content Identification (HASOC)} described by \citet{mandla2021overview}, is part of the Hate Speech and Offensive Content Identification track, consists of 17,657 tweets in Hindi, German, and English, annotated for hate speech and offensive content with three classification levels: presence of hate or offensive content, type of offense (hate, offensive, or profane), and the nature of the insult (targeted or untargeted).

\noindent \textbf{Call me sexist, but (CMS)} described by \citet{samory2021call}, consists of 6,325 instances drawn from Twitter and includes a comprehensive annotation for expressions of sexism, where each entry is independently labeled by five coders into categories based on content and phrasing, such as sexist, maybe-sexist, civil, and uncivil.

\noindent It's important to mention many datasets such as \textbf{AIRAIT} \cite{waseem-hovy-2016-hateful}, only their tweet IDs and their labels were publicly available. Regrettably, a substantial portion of tweets (>= 40\% positive datas) were inaccessible through the X API.

\subsubsection{Data Processing and Splits}\label{dataprocessing}
Our data preprocessing pipeline prioritizes identifying harmful content in English. Thus, we exclude any tweets in languages other than English and verify language using a Roberta-based language identifier. Additionally, to prevent overlap between similar datasets, we remove near-duplicated entries through normalization, ignoring duplicate entries and removing URLs and mentions. The majority of collected datasets focus on hate speech, and in some cases, offensive speech, which we consider harmful. For datasets with a binary hate classification task or a more detailed classification like CMS, where all "hate" subclasses are treated as one, we categorize them as harmful. Datasets focusing on specific types of hate speech, such as sexism, are also considered harmful. Moreover, datasets containing offensive speech are likewise classified as harmful in our case.
Finally, in datasets like AHSD and HASOC, where a distinction between hate, offensive, or profound speech exists, we include entries labeled as hate speech or offensive, considering them as harmful. Entries labeled as normal or not-hateful are categorized as not harmful.

Table  \ref{tab:dataset_breakdown} shows the final numbers of data and classes after processing. Datasets divided into training, validation, and testing sets in a ratio of 7:1:2, respectively.
\begin{table}[h]
    \centering

    \begin{tabular}{ccc}
    \hline\hline
    Dataset & \textbf{Harmful } & \textbf{Non-Harmful} \\
    \hline\hline
    \textbf{AHSD} & 1200* & 4081 \\
    
    \textbf{HTPO} & 351 & 2508 \\
    
    \textbf{HASOC} & 1200* & 4292 \\
    
    \textbf{CMS} & 1203 & 4000* \\
    \hline\hline
    \end{tabular}
    \captionsetup{justification=centering}
    \caption{Dataset Breakdown: Harmful vs. Nonharmful Content Numbers in AHSD, HTPO, HASOC, CMS. * indicates that dataset sizes have been downsized to simulate reality low-resources setting.}
    \label{tab:dataset_breakdown}
\end{table}

\begin{table*}[ht] 
\centering
\scalebox{0.83}{
\begin{tabular}{@{}ccccccccccc@{}}
\toprule
\multicolumn{2}{l}{\multirow{3}{*}{Methods/Seed Count}} & \multicolumn{2}{c}{AHSD} & \multicolumn{2}{c}{HTPO} & \multicolumn{2}{c}{HASOC} & \multicolumn{2}{c}{CMS} \\ 
\cmidrule(l){3-4} \cmidrule(l){5-6} \cmidrule(l){7-8} \cmidrule(l){9-10}
& & BERT & RoBERTa & BERT & RoBERTa & BERT & RoBERTa & BERT & RoBERTa \\ 
\addlinespace
\midrule

\multirow{1}{*}{\centering Gold Labels} & - & 86.0 & 88.0 & \textbf{74.0} & 69.0 & 58.2 & \textbf{57.3} & 85.2 & 85.4 \\
\addlinespace
\midrule
\multirow{4}{*}{In-context Learning w/ Seed} & 50 & 34.9 & 35.2& 23.5 &  22.8 & 8.1 & 10.4 & 34.4 & 37.5\\
& 100 & 47.8 & 49.5 & 48.2 & 46.3 & 17.9 & 20.5 & 49.3 & 52.0  \\
& 150 & 65.2 & 68.1 & 54.6 & 55.1 &21.4 &21.4 &58.9 & 57.4 \\
& 200 & 63.2 & 68.0 & 59.2 & 58.7 & 32.6 & 31.1 &63.8 & 66.2 \\
\addlinespace
\midrule
\multirow{4}{*}{ToxiCraft w/ Seed (Ours) } & 50 & 62.4 &63.7 & 40.6 & 38.9 & 15.2 & 16.0 & 53.2 & 55.9 \\
& 100 & 74.7 &73.9&47.8& 44.3 & 27.3&28.2& 75.6&70.4 \\
& 150 & 85.5 &87.1 & 58.0 &59.3 & 44.1 &43.6 &\underline{86.0} & 84.8 \\
& 200 & \underline{\textbf{89.0}} & \underline{\textbf{88.4}} & 70.3 & \underline{\textbf{70.5}} & \underline{\textbf{58.9}}& 56.1 &\underline{\textbf{86.3}}&\underline{\textbf{86.5}} \\
\bottomrule
\end{tabular}
}
\captionsetup{justification=centering}
\caption{Comparison of Model Performances Across Different Databases and Data Generation Methods. The comparison parameter is MacroF1. Results outperformed training on gold label are underlined while the best performance on each dataset are bolded. Some baseline results come from \citet{antypas-camacho-collados-2023-robust}.}
\label{mainres}
\end{table*}

\subsection{Baselines}

We contrast our method with various baseline approaches, including those that utilize LLM-powered data augmentation techniques.

\noindent \textbf{All Gold Labels}: This method utilizes the entire gold-labeled training dataset for training the models. It serves as a baseline by providing comprehensive data exposure, which allows the models to learn from a complete range of examples in the dataset.

\noindent \textbf{Seed Gold Labels with In-context Learning}: This approach expands the seed data by generating new data instances from the seed data samples, where samples for in-context learning and target-context selection are randomly picked. It tests the model's ability to generalize from an enhanced but limited dataset, simulating training under resource-constrained conditions.

\noindent \textbf{ToxiCraft}:  Our proposed method that enhances seed data with synthetically generated data, expanding the training dataset by adding new examples that mirror the characteristics of the seed data. This method not only expands the quantity of available training data but also diversifies the types of training examples. It aims to overcome data scarcity and improve model robustness by broadening the training scenarios. 

\subsection{Implementation Details}

We employed GPT-4 \cite{openai2024gpt4} as our agent model. For generations, in our experiment, we set N to be 1000, K to be 100, and M to be 3. For smaller-scale classification tasks, we utilized two language models of a reasonable size: BERT-base \cite{devlin2019bert} and RoBERTa-base \cite{liu2019roberta} as downstream models. The implementations provided by Hugging Face \cite{wolf-etal-2020-transformers} were utilized for training and evaluating all language models. The fine-tuning parameters are listed in \ref{sec:appendix2}, and we used macro-F1 to report the evaluation scores, ensuring consistency across models and tasks.

\section{Experimental Results}

\noindent\textbf{Main Results}
We conduct experiments on two different data augmentation scenarios and report the results of training data augmentation in Table \ref{mainres} and the LLM generated successful rate results in Table \ref{tab:success_rate}. As shown in table, using the complete gold-labeled dataset, generally sets a high benchmark, notably on the AHSD and CMS datasets, indicating that comprehensive data access typically results in better model performance. Conversely, `In-context Learning with Seed' shows a performance increase with larger seed sizes, but significantly underperforms with smaller seeds (50, 100), indicating challenges in training models with limited data. 
Our Toxicarft substantially outperforms or comparable with all baselines across different settings demonstrating the effectiveness of our approach. Notably, our approach, which enhances seed data with synthetically generated labels, demonstrates substantial performance gains, especially with larger seeds (150, 200). These results approach or even surpass those from training with all gold labels, highlighting the potential of synthetic data to effectively supplement training datasets and enhance learning outcomes.
The lower performance across all methods in the HTPO dataset, which focuses on political harmful content, suggests that such content poses additional complexities and nuances that are challenging for models to learn. Our framework not only improves as more seed data is available but also illustrates that well-generated synthetic data can serve as a robust tool for augmenting training datasets, thus potentially reducing reliance on extensive manually labeled data. In term of models, RoBERTa generally outperforms BERT, especially in higher data regimes (e.g., "ToxiCraft with Seed Data" at seed 200 across multiple datasets). This could be attributed to RoBERTa's more robust pre-training, which might be better at handling the nuances introduced by synthetic data augmentation.

\begin{table}[t]
\centering

\begin{tabular}{@{}lcccc@{}}
\toprule
Dataset & AHSD & HTPO & HASOC & CMS \\ 
\midrule
Target Size & 1000 & 1000 & 1000 & 1000\\
Success (\%) & 78.6 & 52.2 & 75.2 & 69.1 \\

\bottomrule
\end{tabular}
\captionsetup{justification=centering}
\caption{Target Size and Success Rate of GPT-4 Generated Synthetic Data Across Datasets}
\label{tab:success_rate}
\end{table}

\noindent \textbf{Analysis on Generated Success Rate}
The high success rate of 78.6\% for the AHSD dataset suggests that GPT-4 excels in generating data that closely matches the characteristics and complexities of this particular dataset. It's important to note that both the success rate and the macroF1 generated from a small amount of seed data for AHSD are high, the possibility of a data breach cannot be ruled out. Conversely, the lower success rate of 52.2\% for the HTPO dataset, which focuses on political harmful content, highlights the challenges faced in synthetic data generation when dealing with nuanced and sensitive content. It's possible that the GPT series may have been specifically tuned for political topics, which could explain the difficulties encountered in generating synthetic data for this dataset.

\begin{figure}[h]

  \centerline{\includegraphics[width=0.5\textwidth]{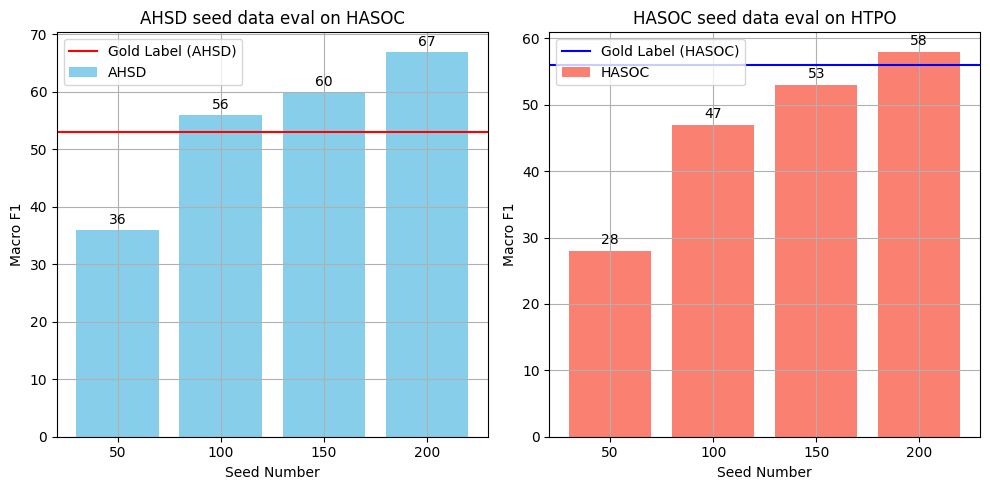}}

  \caption{Harmful detection approaches}
  \label{fig:f3}
\end{figure}

\noindent \textbf{Analysis on Robustness}
Figure \ref{fig:f3} reveals ToxiCraft's robust ability to generate synthetic data that enhances model generalization across datasets. In the AHSD to HASOC evaluation, we observe that as the seed size increases, the performance of models trained on ToxiCraft synthetic data rises significantly, peaking at a Macro F1 score of 67 for the 200 seed size. This surpasses the gold label performance for the AHSD dataset, which stands at a Macro F1 score of 54. Similarly, for the HASOC to HTPO evaluation, the upward trend continues with the performance of the 200 seed size reaching a Macro F1 score of 58, again outperforming the gold label performance, benchmarked at 56. These results not only demonstrate ToxiCraft's capacity to produce high-quality synthetic data that captures the complexity of harmful content but also its effectiveness in adapting to different datasets. The ability to exceed gold label performance suggests that ToxiCraft could potentially reduce the need for extensive labeled datasets, particularly in specialized or sensitive areas like political content, where data acquisition is challenging.

\section{Ablation Study}

In this section, we present the results of our ablation study, which focuses on three key aspects: the choice of model, the impact of different modules, and the robustness of our framework under different settings. These experiments were conducted on the AHSD dataset, HTPO dataset, with 200 seed data and BERT as the downstream model.

\subsection{Study on Backbone Model}
In the early stages of our research, we experimented with several open-source models, including LLaMA2 13B, LLaMA3 8B and Flan-T5-XXL, as alternatives to GPT-4. However, the success rates and overall performance of these models were significantly lower than those achieved by GPT-4.

Table~\ref{tab:model_comparison} summarizes the success rate and macro-F1 scores for each model. AHSD Dataset with 100 seed data is listed in \ref{sec:appendix3}.
GPT-4 clearly outperforms the alternatives, which justifies its selection as the preferred model for our experiments.

\begin{table}[h]
    \centering
    \begin{tabular}{lcc}
    \hline
    \textbf{Model} & \textbf{Success Rate (\%)} & \textbf{Macro-F1} \\
    \hline
    GPT-4 & 78 & 89.0 \\
    LLaMA2 13B & 32 & 38.7 \\
    Flan-T5-XXL & 27 & 34.1 \\
    LLaMA3 8B & 22 & 62.5 \\
    \hline
    \end{tabular}
    \caption{Comparison of models on the AHSD dataset with 200 seed data and BERT as the downstream model.}
    \label{tab:model_comparison}
\end{table}

\subsection{Study on Modules }
To assess the contributions of individual components in our framework, we conducted ablation studies by systematically removing key modules. The results, displayed in Table~\ref{tab:ablation_study}, highlight the significance of each module in model performance.

\noindent \textbf{ToxiCraft Prompt:} Removing the manually-set indicators in the ToxiCraft Prompt caused a notable decrease in macro-F1 score, from 89.0 to 85.2 on the AHSD dataset, and from 70.3 to 58.5 on the HTPO dataset, underscoring the importance of prompt tuning in the data generation process.

\noindent \textbf{CAE:} Excluding the CAE module reduced performance to 84.3 macro-F1 on the AHSD dataset and 63.8 on the HTPO dataset, demonstrating its crucial role in capturing contextual information.

\noindent \textbf{TSR:} Removing the TSR module resulted in the lowest performance (83.6 macro-F1 for AHSD, 65.1 for HTPO), further emphasizing its importance in improving model robustness and generalization.

\begin{table}[h]
    \centering
    \begin{tabular}{lcc}
    \hline
    \textbf{Method} & \textbf{AHSD} & \textbf{HTPO} \\
    \hline
    Full model & 89.0 & 70.3 \\
    w/o ToxiCraft Prompt* & 85.2 & 58.5 \\
    w/o CAE & 84.3 & 63.8 \\
    w/o TSR & 83.6 & 65.1 \\
    \hline
    \end{tabular}
    \caption{Ablation study results on both the AHSD and HTPO datasets using MacroF1. * indicates ToxiCraft Prompt with manually set indicators.}
    \label{tab:ablation_study}
\end{table}

\subsection{Study of the Cross-Dataset Benefits of Modules }

To further validate the robustness of our framework, we evaluated its performance on the HASOC dataset, using the AHSD dataset for data generation. As shown in Table~\ref{tab:robustness_study}, the full ToxiCraft framework exhibited the highest macro-F1 score (67.3), confirming its effectiveness in generating robust training data across datasets. When the manually set indicators in the ToxiCraft Prompt were removed, the macro-F1 score dropped significantly to 49.5, further underscoring the importance of these indicators in enhancing data diversity and performance. The exclusion of the CAE module resulted in a lower macro-F1 score of 50.3. Similarly, the absence of TSR led to a reduction in performance (57.4 macro-F1). In comparison, using only gold label data yielded a macro-F1 score of 54.7, showing the advantage of our method.

\begin{table}[h]
    \centering
    \begin{tabular}{lc}
    \hline
    \textbf{Method} & \textbf{Macro-F1} \\
    \hline
    Gold label & 54.7 \\
    ToxiCraft & 67.3 \\
    ToxiCraft Prompt w/ * & 58.6 \\
    
    ToxiCraft Prompt w/o * & 49.5 \\
    w/o CAE & 50.3\\
    w/o TSR & 57.4\\

    \hline
    \end{tabular}
    \caption{Robustness study on the HASOC dataset using data generated from the AHSD dataset, * means manually set indicators.}
    \label{tab:robustness_study}
\end{table}

\section{Managing Risks of Generating Hateful Content}

The use of large language models (LLMs) to generate hateful content presents significant ethical concerns, particularly around amplifying harmful biases. To mitigate these risks, we implemented strict content review processes to filter inappropriate outputs and limited the scope of generation to controlled experimental environments. We carefully ensured that synthetic data remained focused on research objectives. All experiments adhered to ethical guidelines, underwent reviews, and complied with responsible AI standards. The datasets generated were made available with clear warnings to promote transparency and encourage responsible use in future research. We will actively collaborate
with the community to monitor, responding swiftly to any indications of misuse.

\section{Conclusion}
In this study, we introduced ToxiCraft, a framework that effectively enhances data augmentation for harmful content detection tasks in low-resource settings. ToxiCraft utilizes Large Language Models to synthetically expand seed datasets, overcoming the diversity and volume limitations of conventional augmentation methods. Our extensive evaluations demonstrate that ToxiCraft significantly improves model robustness, outperforming or closed to baselines trained with gold label datasets. This work contributes to the ongoing efforts to develop data-efficient and adaptable models for sensitive content moderation, and sets a foundation for future research on leveraging synthetic data generation in various domains.

\section{Future Work}
Future research for ToxiCraft will focus on three main objectives: 1) enhancing multilingual capabilities, 2) refining the seed data selection process, and 3) reducing reliance on high-cost LLMs. For multilingual data generation, future efforts will explore translating harmful content from English into other languages while preserving both local nuances and the target language characteristics. In terms of optimizing seed data selection, the focus will be on refining the methodology to identify the most diverse and representative examples within a dataset. One proposed approach is to analyze word embeddings and select seed data points whose vectors are the least similar, ensuring broad coverage of the data space. Lastly, cost-effective model exploration will involve identifying more affordable and transparent alternatives to GPT-4, such as the Mistral model, for generating high-quality synthetic samples of harmful content \cite{hui2024openllms}.
\clearpage

\section{Limitations}
In our investigation of the ToxiCraft framework, we have recognized a few limitations that warrant further attention: (1) While our approach, which centers around GPT-4, has shown proficiency in generating synthetic data for harmful content detection, its performance on niche or underrepresented content types is yet to be fully understood. The adaptability of ToxiCraft to a wider array of nuanced domains remains an area for exploration. (2) The efficacy of ToxiCraft relies partly on the availability of initial seed data of high quality. Obtaining such data can pose challenges, particularly in highly specialized or sensitive contexts. Future work will need to address strategies for seed data selection in scenarios where gold standard labels are scarce or non-existent. (3) The utilization of models like GPT-4 brings about concerns related to accessibility and reproducibility. The proprietary nature of such models and potential licensing restrictions may limit widespread adoption and independent verification of the results.
\section{Ethical Consideration}
The generation of synthetic data for harmful content detection necessitates careful ethical considerations. It involves handling potentially sensitive or offensive material, and there is a responsibility to ensure that such data does not perpetuate harm or bias. Rigorous validation processes and ethical oversight are essential to prevent the reinforcement of such biases in synthetic data. Collaboration with subject matter experts and ethicists will be critical to navigate these challenges effectively and responsibly in future iterations of ToxiCraft.

\bibliography{custom}
\clearpage
\appendix

\section{Pseudocode for ToxiCraft}
\label{sec:appendix1}

\begin{algorithm}
\caption{ToxiCraft Data Augmentation}
\begin{algorithmic}[1]

\Procedure{ToxiCraft}{$D, D_h, n$}
    \State $S \gets \text{SelectSeedData}(D_h, n)$ \Comment{Select seed data from harmful examples}
    \State $D' \gets D$ \Comment{Initialize augmented dataset with original data}
    \For{each $(x, y) \in S$}
        \State $\epsilon \sim \mathcal{P}(\epsilon)$ \Comment{Sample noise vector}
        \State $x' \gets \Phi(\text{LLM}, x, \epsilon)$ \Comment{Generate synthetic instance}
        \State $D' \gets D' \cup \{(x', y)\}$ \Comment{Add synthetic instance to dataset}
    \EndFor
    \State \textbf{return} $D'$
\EndProcedure

\Function{SelectSeedData}{$D_h, n$}
    \State $indices \gets \text{RandomIndices}(D_h, n)$ \Comment{Get random indices for seed selection}
    \State \textbf{return} $\{(x^{(i)}, y^{(i)}) : i \in indices\}$
\EndFunction

\Function{RandomIndices}{$D_h, n$}
    \State $count \gets \text{size of } D_h$
    \State \textbf{return} $n$ random unique indices from $1$ to $count$
\EndFunction

\end{algorithmic}
\end{algorithm}

\section{Training parameter}
\label{sec:appendix2}
We fine-tuned the learning rate, warm-up steps, number of epochs, batch size, and other hyperparameters for each model. For BERT, we use learning rate=$1.8282\times10^{-5}$, epoch=3, batch=4, warm-up step=30.And for RoBERTa, we use learning rate=$1.1856\times10^{-5}$, epoch=3,  batch=4, warm-up step=30. All fine-tuning experiments were conducted using A40 and GTX 4080 Super GPU, and the results are reported in terms of macro-F1 score. 

\newpage

\section{AHSD Dataset with 100 seed data Success rate}

Table \ref{tab:7} presents the performance of different models on the AHSD dataset, using 100 seed data and evaluated with BERT as the downstream model.

\label{sec:appendix3}
\begin{table}[h]
    \centering
    \begin{tabular}{lcc}
    \hline
    \textbf{Model} & \textbf{Success Rate (\%)} & \textbf{Macro-F1} \\
    \hline
    GPT-4 & 76 & 74.7 \\
    LLaMA2 13B & 42 & 10.8 \\
    Flan-T5-XXL & 33 & 16.1 \\
    \hline
    \end{tabular}
    \caption{Comparison of models on the AHSD dataset with 100 seed data and BERT as the downstream model.}
    \label{tab:7}
\end{table}

\section{Example of ToxiCraft framework}

\begin{figure*}[ht]
  \begin{center}
  \includegraphics[width=1\textwidth]{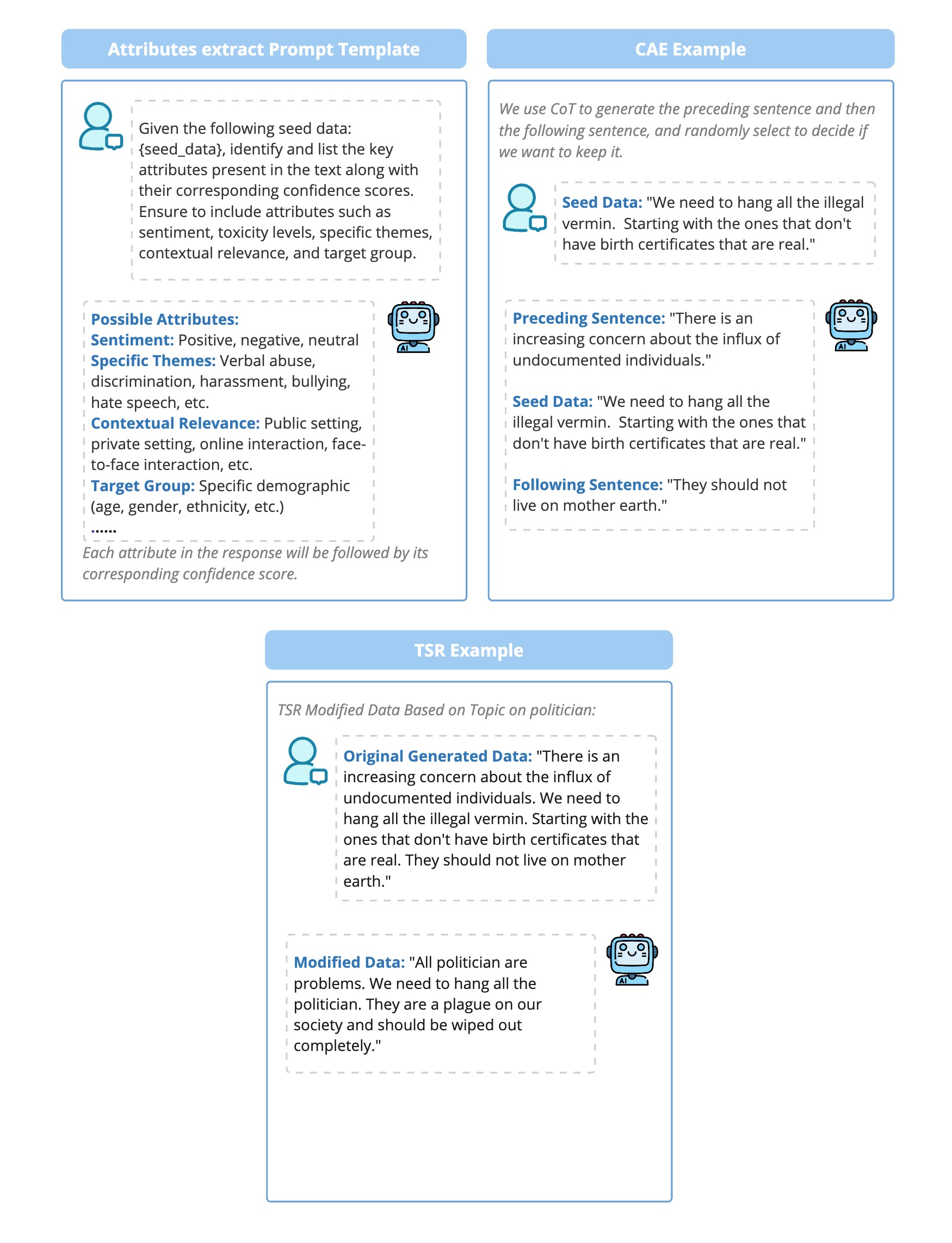}
  \end{center}
  \captionsetup{justification=centering}
  \caption{Our \textbf{ToxiCraft} Framework example}
  \label{fig:example}
\end{figure*}

In Figure \ref{fig:example}, we provide a example of our ToxiCraft to help better understand.
\label{sec:appendix4}

\end{document}